\documentclass{article}
\usepackage{spconf,amsmath,graphicx}
\usepackage{enumitem}
\setlist{nosep, leftmargin=14pt}
\usepackage[table]{xcolor} 
\usepackage{mwe} 
\usepackage{color}
\usepackage[numbers]{natbib}
\usepackage[skip=0.4pt]{caption}
\usepackage{hyperref}
\title{TractoGPT: A GPT architecture for White Matter Tract Segmentation}

\name{
    \begin{tabular}{c}
    Anoushkrit Goel$^{1}$, Simroop Singh$^{1}$, Ankita Joshi$^{1}$, Ranjeet Ranjan Jha$^{2}$,\\ Chirag Ahuja$^{3}$, Aditya Nigam$^{1}$, Arnav Bhavsar$^{1}$
    \end{tabular}
}
\address{
    $^1$School of Computing and Electrical Engineering, Indian Institute of Technology Mandi \\
    $^2$Department of Mathematics, Indian Institute of Technology Patna \\
    $^3$Post Graduate Institute of Medical Education Research, Chandigarh \\
}

\begin{document}
\maketitle
\begin{abstract}
White matter bundle segmentation is crucial for studying brain structural connectivity, neurosurgical planning, and neurological disorders. White Matter Segmentation remains challenging due to structural similarity in streamlines, subject variability, symmetry in 2 hemispheres, etc. To address these challenges, we propose \textbf{TractoGPT}, a GPT-based architecture trained on streamline, cluster, and fusion data representations separately. TractoGPT is a fully-automatic method that generalizes across datasets and retains shape information of the white matter bundles.
Experiments also show that \textbf{TractoGPT} outperforms state-of-the-art methods on average DICE, Overlap and Overreach scores. We use TractoInferno and 105HCP datasets and validate generalization across datasets.

\end{abstract}
\vspace{-0.3cm}
\section{Introduction} \label{sec:intro}
Fiber tract segmentation is a pivotal process in Neuroimaging, enabling detailed analysis of White Matter connectivity through Diffusion Magnetic Resonance Imaging (dMRI).
Tractography traces the anisotropic diffusion of water molecules along neural pathways, yielding three-dimensional streamlines that represent white matter fiber bundles. 
These streamlines are grouped into specific anatomical tracts, providing crucial insights into brain connectivity and function, essential for understanding development, ageing, and neurological conditions \cite{basser1994mr}.

With recent advancements, fiber tract segmentation methods can be broadly categorized into classical and deep learning techniques. Classical methods, such as \textit{QuickBundles} and \textit{QuickBundlesX}\cite{garyfallidis2012quickbundles} cluster streamlines into bundles, Fast Streamline Search \cite{st2022fast} searches similar streamlines accurately,  RecoBundles\cite{garyfallidis2018recognition} recognizes model bundles, utilizing distance metrics like Mean Direct-Flip Distance (MDF)\cite{garyfallidis2012quickbundles}. Deep Learning techniques like TractSeg\cite{wasserthal2018tractseg} employs Convolutional Neural Networks (CNNs) across multiple MRI slices, DeepWMA \cite{zhang2020deep} utilizes novel fiber descriptors, 
FINTA \cite{legarreta2021filtering} does filtering in embedding space, and FIESTA \cite{dumais2023fiesta} further improves the FINTA by employing FINTA-multibundle to segment and GESTA-GMM \cite{legarreta2023generative} to fill bundles to meet semi-automatically calibrated bundle-specific thresholds.
Recent tract segmentation studies have also explored point cloud networks \cite{xue2023tractcloud, goel2024tractoembed}, but most methods either require registration, ATLAS, filtering or calibration for thresholds.
In this paper, 
\begin{itemize} 
\item We introduce \textbf{TractoGPT}, a novel, fully-automatic, \textbf{registration-free} white matter segmentation network inspired by the GPT architecture.
\item In addition, we introduce a\\ \textbf{Fusion} Data Representation which enhances representation for tractography streamline data for downstream segmentation task. 
\item TractoGPT also \textbf{generalizes across datasets} and retains shape information of major White Matter Bundles. 
\end{itemize}
\vspace{-0.3cm}
\section{Methodology} \label{sec:method}
\begin{figure*}[hbt!]
    \centering
    \includegraphics[width=1\linewidth]{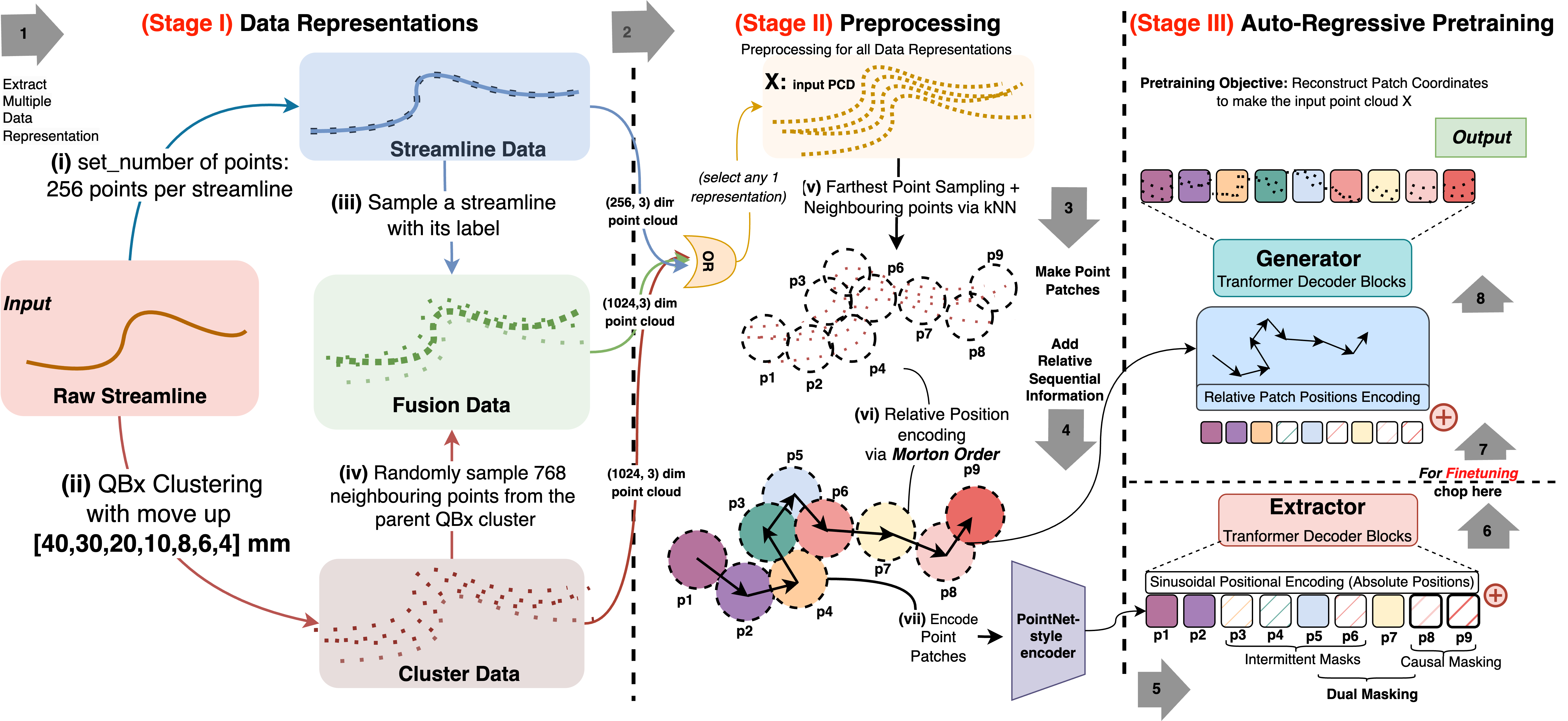}
    \caption{\textbf{TractoGPT Architecture:} \textbf{(Stage I)} Raw Streamline  undergoes preprocessing \textit{(\textbf{i, ii, iii, iv})} to give us 3 different data representations (Section \ref{sec:preprocess}). \textbf{(Stage II)}  From 3 different point cloud arrays, any one can be chosen to train TractoGPT, Extracted Point Cloud undergoes \textit{FPS (Farthest Point Sampling)} to give total $P$ center points (Absolute Positions), used to sample a total of $K$ nearest neighbors using \textit{kNN} (\textbf{v}), creating point cloud patches. \textbf{(Stage III)} Point patches get sequence using Morton order (Relative Positions) (\textbf{vi}), and a PointNet-style encoder gives embedding for each patch as tokens.}
    \label{fig:landing}
\end{figure*}
\subsection{dMRI Datasets and Tractography}
\label{sec:data}
\vspace{-0.3cm}


\setlength{\arrayrulewidth}{0.1pt}
\begin{table}[htb!]
    \centering
    \begin{tabular}{cp{0.7cm}p{1.2cm}}
    \hline
         \textbf{Dataset} & \textbf{Class} & \textbf{Subjects} \\
         \hline
         \textbf{105 HCP} \\(Human Connectome Project) \cite{van2013wu, wasserthal2018tractseg}& 72 & 80:15:10\\
         \hline
         \textbf{TractoInferno} \cite{poulin2022tractoinferno} & 32 & 198:58:28\\
         \hline
    \end{tabular}
    \caption{Under the \textit{Subjects} column, denote subjectwise data splits of\textbf{ train:val:test} used in TractoGPT. HCP and TractoInferno Datasets are publicly available.}
    \label{tab:datasets}
\end{table}
We use datasets mentioned in Table \ref{tab:datasets} where \textbf{TractoInferno}\cite{poulin2022tractoinferno} is a silver-standard dataset created by ensembling 4 tracking methods and RecoBundlesX \cite{garyfallidis2018recognition} to generate ground truth streamlines  and recognize bundles respectively, yielding 32 classes using an ATLAS. 
Whereas \textbf{105 HCP} \cite{wasserthal2018tractseg} dataset is created from raw HCP data using MRtrix3 and TractSeg, and contains 72 classes.
\vspace{-0.3cm}
\subsection{Data Representations} \label{sec:preprocess}
Process of Whole Brain Tractography (WBT) yields streamlines, i.e. variable sequence of 3D coordinates which are input to TractoGPT along with label of each streamline.

To embed a richer understanding of tractography data to the model, we propose 3 data representations, (refer Fig. \ref{fig:landing}, Stage I), either of which can be used as the Training Data. Due to time efficiency, and bounded parent clusters, we use QuickBundlesX for finding neighbouring streamlines.
\begin{itemize}
    \item \textbf{Streamline}: Streamlines can be of variable length, hence we bicubic interpolate streamlines of (n,3) dimension to get a (256,3) dimensional array, as shown in Fig.\ref{fig:landing} Stage I.
    \item \textbf{Cluster}: We provide streamlines with relative location information by sampling clusters of streamlines resembling parent bundles. 
    We modify QuickBundlesX (QBx) clustering to devise, \textit{QBx Clustering with move up}. Here, clusters are initially formed at hierarchical thresholds of [40,30,20,10,8,6,4] mm, but only the finest three levels (4 mm, 6 mm, and 8 mm) are used for training.
    To ensure the quality of the cluster, we sample a minimum threshold of 10 streamlines per cluster.
    Beginning from the finest level (4 mm), if a cluster lacks the required number of streamlines, then the method moves up to the next coarser level (6 mm, \textit{then} 8 mm) to use clusters sampled in new radius.
    Clusters formed above 8mm are discarded to avoid presence of multiple classes in coarser clusters.
    
    After a cluster is formed, 1024 random points are selected from a cluster (group of streamlines) to create a point cloud with a shape of $(1024,3)$, refer (ii) in Fig.\ref{fig:landing}. 
    
    \item \textbf{Fusion}: Fusion Data is a fusion of streamline and cluster data allowing the amalgamation of both representations. For fusion data, 256 points are sampled from the interpolated streamline of interest, and the rest 768 points are sampled from the non-interpolated neighboring streamlines to make a 1024-dimensional array (see (iii), (iv) in Fig.\ref{fig:landing}).
\end{itemize}


\begin{figure*}[hbt!]
    \centering
    \includegraphics[width=1\textwidth]{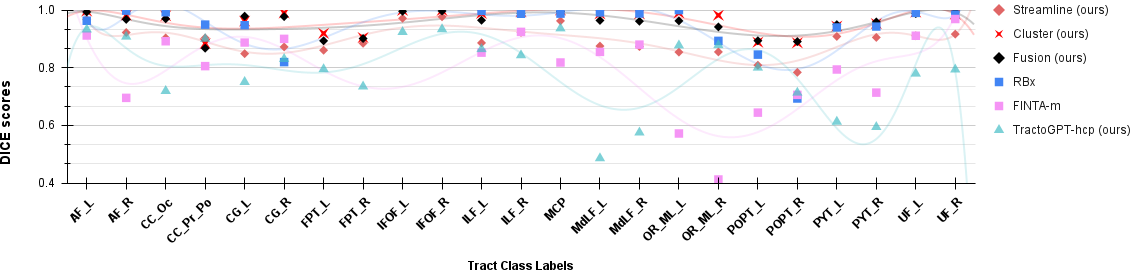}
    \caption{\textbf{Voxel DICE scores for class-wise comparison} across \textit{FINTA-m} \textit{RecoBundlesX}, \textit{TractoGPT-hcp}, \textit{TractoGPT} methods on TractoInferno test dataset. \textbf{Class-wise Ablation study} of TractoGPT are across [streamline, cluster, fusion] data representations. Here \textit{TractoGPT-hcp} results are shown for \textbf{dataset generalization}, which is trained on HCP and tested on TractoInferno}
    \label{graph}
\end{figure*}
\vspace{-0.3cm}
\subsection{Tokenization}\label{sec:token}
A group of points yields regional information on the shape of the point cloud. We create point patches to embed regional information in tokens using an encoder network.

Point patches are obtained through \textbf{FPS-kNN} (Farthest Point Sampling \& k-Nearest Neighbors) where the farthest points are treated as centroids to make patches by sampling K neighbors via kNN. For GPT architecture, sequential information is required among tokens which are extracted using sorted Morton Order (or Z-order curve) on encoded $K$ center points (1-dimensional array), and this Relative Positional Encoding is passed to the Generator part of TractoGPT (see Fig. \ref{fig:landing})\cite{chen2024pointgpt}.
The coordinates of each point are normalized relative to its center point before they are fed to the PointNet-style encoder \cite{qi2017pointnet}, giving a latent representation per patch to make tokens, along with the Morton order sequence. 
While setting patch configuration, a patch size of 32 points for [Fusion, Cluster] and 8 points for Streamline. 
Number of patches is set to 64 for all representations to ensure overlapping patches.
\vspace{-0.4cm}
\subsection{TractoGPT Model}
In TractoGPT, we employ an architecture consisting of Extractor and Generator which are essentially stacked transformer decoder blocks \cite{chen2024pointgpt, yu2022point}. The model undergoes autoregressive pretraining separately on all data representations  (refer Fig. \ref{fig:landing}).
The overall pre-training objective of our model is to reconstruct input point cloud patches using sequence of tokens extracted from the Tokenization process above.
In pre-training we use \textit{dual masking strategy} which does intermittent masking on top of causal masking inhibiting model to overfit on the input point cloud data.
Causal masking is usually used for Next Token Prediction tasks, and intermittent masking can be understood as a proportion of masked preceding tokens attending to each unmasked preceding token. 

Due to random transformations and shuffling of points while training, order of points is not preserved, leading to ambiguity in predicting consecutive patches. To mitigate this ambiguity, Extractor takes Sinusoidal Positional encoded tokens with absolute normalized positions of center points, and the Generator incorporates directions as Relative Patch Positions Encoding (refer Fig. \ref{fig:landing} Stage II and III), without disclosing the locations of masked patches or the overall shapes of the point clouds. The \textbf{Prediction head} of the Generator is designed to predict subsequent point patches in the coordinate space. The Generator comprises of 2 FCNN layers with (ReLU) activation functions, being shallower than the Extractor.


\begin{figure}[hbt!]
    \centering
    \includegraphics[width=1\linewidth]{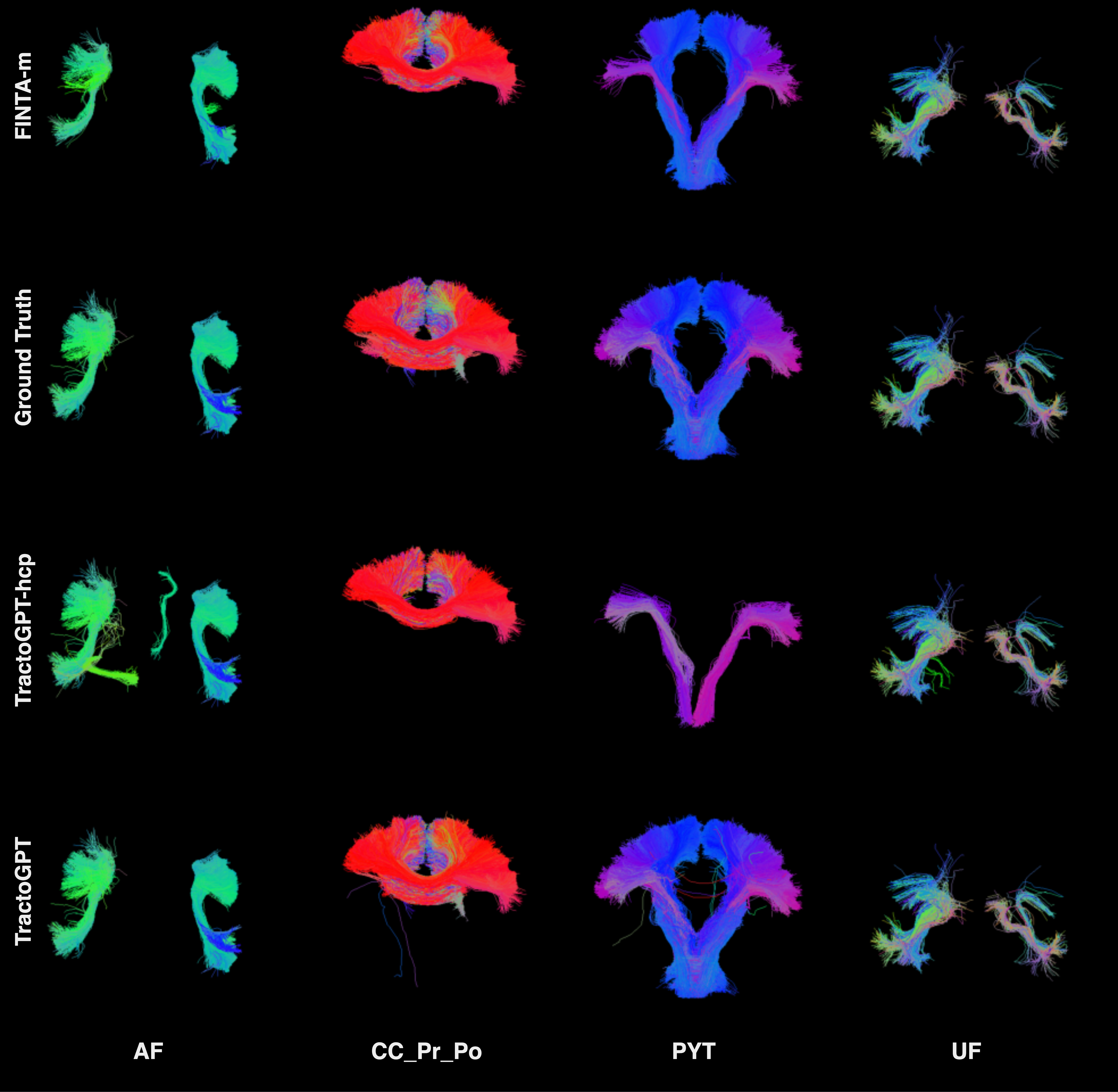}
    \caption{Visualisation of major bundles, tested on \textit{sub-1006}}
    \label{plots}
\end{figure}
\vspace{-0.3cm}
\subsection{Model Training and Testing}
A consistent number of streamlines per class across all training subjects is used to mitigate imbalance for training TractoGPT (for example: we use 500 streamlines per subject per class on a single streamline data representation which can vary based on the choice of data representation).
But while testing, all streamlines of a test subject are classified, without leaving a single streamline behind.
In \textbf{TractoGPT} training and testing strategy, we implement a comprehensive approach that includes pretraining, fine-tuning, and testing.
\textbf{Pretraining} involves reconstruction of patch coordinates using a 50:50 Chamfer Loss (L1 and L2 norm) without labels, optimized with AdamW (weight decay of 0.05) and a cosine learning rate scheduler starting at 0.0001 learning rate, over a maximum of 150 epochs (converges earlier).
In \textbf{fine-tuning} for classification, we employ Cross Entropy and Chamfer Distance Loss (CDL1 + CDL2) in a ratio of 1:3 respectively.
The overall strategy is designed to optimize the model's performance in classification and reconstruction tasks while leveraging advanced loss functions for effective tractography streamlines understanding.
TractoGPT takes less than 4 days on average to converge (cluster $<$ streamline $<$ fusion) and 12 hours to infer on a single V100 16 GB GPU on TractoInferno Dataset.
\vspace{-0.4cm}
\section{Experiments and Results}

\begin{table}[hbt!]
    \centering
    \begin{tabular}{|c|c|c|c|}
    \hline
    \multicolumn{4}{|c|}{\textbf{Train}: TractoInferno \& \textbf{Test}: TractoInferno} \\
    \hline
    \textbf{Representation} & \textbf{DICE} & \textbf{Overlap} & \textbf{Overreach} \\
    \hline
    Streamline & 0.88$\pm$0.07 & 0.82$\pm$0.11 & \textbf{0.03}$\pm$\textbf{0.04} \\
    \hline
    Cluster & \textbf{0.96}$\pm$\textbf{0.04} & \textbf{0.96}$\pm$\textbf{0.04} & 0.04$\pm$0.06 \\
    \hline
    Fusion & 0.95$\pm$0.04 & 0.94$\pm$0.05 & 0.04$\pm$0.07 \\
    \hline
    \multicolumn{4}{c}{} \\ [-1.5ex] 
    \hline
    \multicolumn{4}{|c|}{\textbf{Train}: HCP \& \textbf{Test}: TractoInferno} \\
    \hline
    \textbf{Representation} & \textbf{DICE} & \textbf{Overlap} & \textbf{Overreach} \\
    \hline
    Streamline & 0.73$\pm$0.1 & 0.68$\pm$0.16 & 0.16$\pm$0.25 \\
    \hline
    Cluster & 0.79$\pm$0.13 & \textbf{0.78}$\pm$\textbf{0.18} & 0.28$\pm$0.5 \\
    \hline
    Fusion & \textbf{0.79}$\pm$\textbf{0.11} & 0.75$\pm$0.15 & \textbf{0.13}$\pm$\textbf{0.22} \\
    \hline
    \end{tabular}
    \caption{\textbf{Ablation Study}: \textit{Average} test results across all test subjects of TractoInferno when trained on either TractoInferno or HCP}
    \label{tab:ablation}
\end{table}

\begin{table}[hbt!]
    \centering
    \begin{tabular}{|c|c|c|c|}
    \hline
   \textbf{Methods}  & \textbf{DICE} & \textbf{Overlap} & \textbf{Overreach} \\
    \hline
   \color{red}{\textbf{TractoGPT}} & \textbf{0.97}$\pm$\textbf{0.05} & 0.96$\pm$0.07 & \textbf{0.03}$\pm$ \textbf{0.05} \\
    \hline
   RBx  &  0.95$\pm$0.07& 0.95$\pm$0.10 & 0.04$\pm$0.09\\
    \hline
   FINTA-m  & 0.79$\pm$0.13 & \textbf{0.99}$\pm$\textbf{0.01} & 0.59$\pm$0.59\\
    \hline
   TractoGPT-hcp  &  0.79$\pm$0.12 & 0.78$\pm$0.18 & 0.28$\pm$0.50\\
    \hline
    \end{tabular}
    \caption{\textbf{Comparative Study}: \textit{Average} scores across tracts/classes for 1 subject \textit{sub-1006} of TractoInferno (along with standard deviations}
    \label{tab:comparative}
\end{table}

We perform rigorous experiments with the current state-of-the-art tractography segmentation models, FINTA-m, RBx across the list of \textbf{23 common tracts} in TractoInferno and HCP dataset. For a fair and consistent comparison across methods we use DICE scores.
We show the Ablation Study, in the Table \ref{tab:ablation}, where we can see Fusion is comparable to the Cluster, proving the efficacy of Fusion Representation at large.
For the comparative study (see \ref{tab:comparative}), we chose TractoGPT in cluster data representation because it performs better than other methods on TractoInfero dataset.
We show tract-wise comparison of TractoGPT [cluster] with other SOTA methods as shown in Figure \ref{graph} where we can see that our method segments all 23 common major bundles with good DICE scores. 
Here, RecoBundlesX outputs are not filtered using dMRIQC
\cite{theaud2022dmriqcpy}.
We also demonstrate \textbf{TractoGPT's generalization capability}, by training TractoGPT on HCP dataset and testing on TractoInferno test data (see resultes in Table\ref{tab:ablation}), we named it \textit{TractoGPT-hcp}. 
TractoGPT-hcp indicates better generalization than FIESTA. FIESTA reports DICE score of $0.74 \pm 0.08$ on a private dataset, Myeloinferno\cite{dumais2023fiesta}, whereas TractoGPT-hcp achieved $0.79 \pm 0.12$ on TractoInferno. 
\vspace{-0.2cm}
\section{Conclusion}
\vspace{-0.2cm}
In this study, we propose TractoGPT, a novel GPT-based architecture for White Matter Tract Segmentation with SOTA results on TractoInferno dataset, proven potential of generalization across datasets, while preserving shape information of White Matter bundles. We introduced Fusion Data which enriches Streamline-only data representation for segmentation.
\vspace{-0.5cm}
\section{Compliance with ethical standards}
\label{sec:ethics}
This research study was conducted retrospectively using human subject data made available in open access by (TractoInferno \cite{poulin2022tractoinferno}, Human Connectome Project \cite{wasserthal2018tractseg}). Ethical approval was not required as confirmed by the license attached with the open access data.

\vspace{-0.4cm}
\section{Acknowledgments}
\label{sec:acknowledgments}
This work was supported IIT Mandi by SERB CORE Research Grant with Project No: CRG/2020/005492
\vspace{-0.1cm}


\setlength{\bibsep}{0pt plus 0.3ex}
\renewcommand{\bibfont}{\small}
\section{References}
\vspace{-1cm}
\bibliography{refs-sq}
\end{document}